\documentclass{article}

\usepackage{microtype}
\usepackage{graphicx}
\usepackage{subfigure}
\usepackage{booktabs} 
\usepackage{listings}
\usepackage[most]{tcolorbox}
\usepackage{cuted}   
\usepackage{xcolor}
\usepackage{tabularx}
\usepackage{longtable}
\usepackage{hyperref}
\usepackage{tcolorbox}
\usepackage{listings}
\usepackage{cuted}
\usepackage{appendix}

\usepackage[accepted]{icml2024}

\usepackage{amsmath}
\usepackage{amssymb}
\usepackage{mathtools}
\usepackage{amsthm}

\usepackage[capitalize,noabbrev]{cleveref}

\theoremstyle{plain}

\theoremstyle{definition}

\theoremstyle{remark}

\usepackage[textsize=tiny]{todonotes}

\icmltitlerunning{TracrBench: Generating Interpretability Testbeds with Large Language Models}

\begin{document}

\twocolumn[
\icmltitle{TracrBench: Generating Interpretability Testbeds with Large Language Models}



\icmlsetsymbol{equal}{*}

\begin{icmlauthorlist}
\icmlauthor{Hannes Thurnherr}{independent}
\icmlauthor{Jérémy Scheurer}{apollo}
\end{icmlauthorlist}

 \icmlaffiliation{apollo}{Apollo Research, London, UK}
 \icmlaffiliation{independent}{Independent, Zurich, CH}

\icmlcorrespondingauthor{Hannes Thurnherr}{hannes.thurnherr@gmail.com}
\icmlcorrespondingauthor{Jérémy Scheurer}{jeremy@apolloresearch.ai}

\icmlkeywords{Testbed, Interpretability, Sandbox}

\vskip 0.3in
]



\printAffiliationsAndNotice{} 

\begin{abstract}
Achieving a mechanistic understanding of transformer-based language models is an open challenge, especially due to their large number of parameters. Moreover, the lack of ground truth mappings between model weights and their functional roles hinders the effective evaluation of interpretability methods, impeding overall progress.
Tracr, a method for generating compiled transformers with inherent ground truth mappings in RASP, has been proposed to address this issue. However, manually creating a large number of models needed for verifying interpretability methods is labour-intensive and time-consuming.
In this work, we present a novel approach for generating interpretability test beds using large language models (LLMs) and introduce  TracrBench, a novel dataset consisting of 121 manually written and LLM-generated, human-validated RASP programs and their corresponding transformer weights. During this process, we evaluate the ability of frontier LLMs to autonomously generate RASP programs and find that this task poses significant challenges. GPT-4-turbo, with a 20-shot prompt and best-of-5 sampling, correctly implements only 57 out of 101 test programs, necessitating the manual implementation of the remaining programs. 
With its 121 samples, TracrBench aims to serve as a valuable testbed for evaluating and comparing interpretability methods. 

\end{abstract}

\section{Introduction}\label{introduction}
Recent advancements in transformer-based language models have led to progress in various natural language processing tasks~\citep{ achiam2023gpt, anthropic2024claude}. However, understanding the internal workings of these models remains challenging~\citep{olah2018building, nanda2023progress, black2022interpreting}, which is problematic since models may generate harmful outputs~\citep{shevlane2023model, Perez2022Red, brundage2018malicious} or harbor other unacceptable failure modes only revealed after deployment~\citep{ngo2022alignment, scheurer2023technical, hubinger2024sleeper}.
Despite various successes in interpretability~\citep{bricken2023towards, conmy2023towards, nanda2023progress, cunningham2023sparse, templeton2024scaling}, developing new interpretability methods remains difficult, partly due to the lack of models with fully understood internals~\citep{casper2023red, eis_vii}, i.e. with ground truth mapping between weights and their functional form. Existing benchmarks for evaluating interpretability methods focus on input-output behavior \citep{casper2024satml, casper2023red, mazeika2022trojan}, human evaluations \citep{templeton2024scaling}, or disentangling attributions of different entities~\citep{huang2024ravel}, rather than the full mechanistic circuits, which hinders the rigorous and fast validation of novel interpretability methods.

\begin{figure*}[t]
    \centering
    \includegraphics[width=0.49\linewidth]{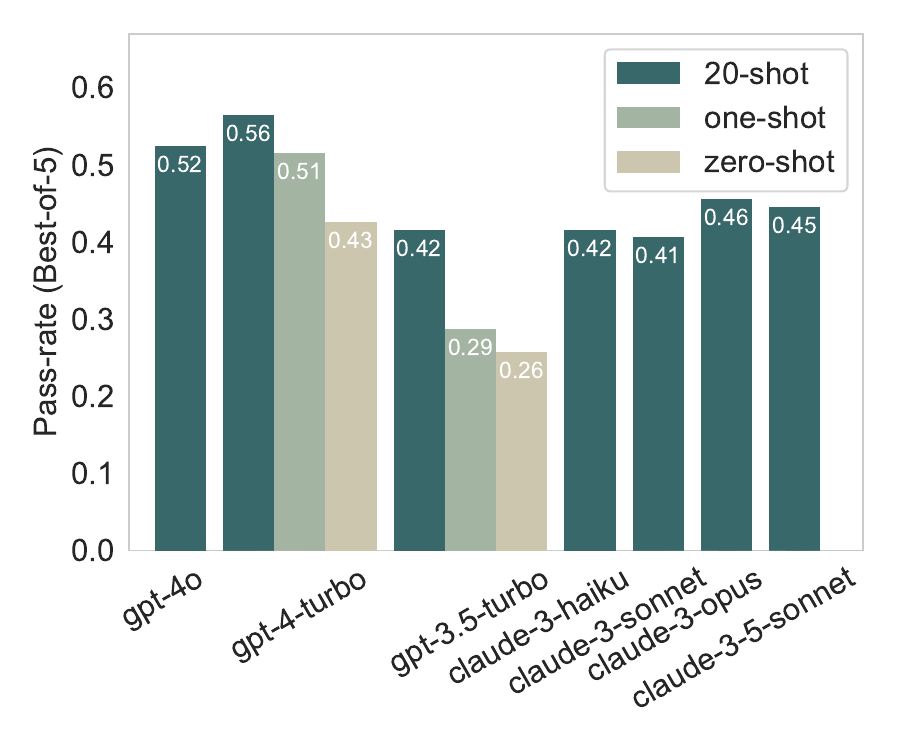}
    \includegraphics[width=0.49\linewidth]{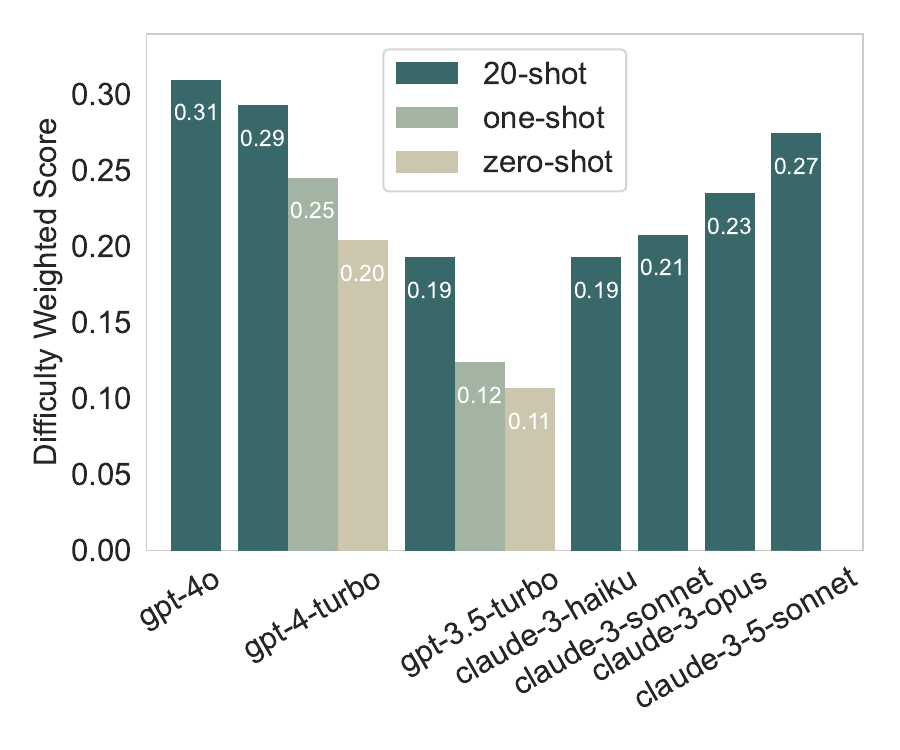}
    \vspace{-20pt}
    \caption{Results on the test set with 101 Tracr programs with pass-rate on the left and a normalized, difficulty-weighted score on the right (maximum score on both metrics is $1.0$). The 20-shot prompt with best-of-5-sampling achieves the best performance to other prompts. gpt-4-turbo-2024-04-09 and gpt-4o models achieve the best performance overall. However, the task is challenging for all models.}
    \label{fig:combined}
\end{figure*}

Restricted Access Sequence Processing Language (RASP) ~\citep{weiss2021thinking} maps the core components of a transformer-encoder, i.e., attention and feed-forward computation, into simple primitives, forming a programming language to model and analyze transformer behavior.
Tracr \citep{lindner2024tracr}, compiles RASP programs into functional transformer weights with a known mapping from weights to their functional form, enabling, the evaluation of interpretability methods \citep{conmy2023towards}.  However, its adoption is limited due to the difficulty of writing RASP programs and the large number of models required to effectively evaluate interpretability methods. 

In this work, we introduce and evaluate a method to automatically generate RASP programs using LLMs and present TracrBench, a novel dataset with 121 LLM generated and, where necessary, manually written RASP programs and their compiled transformers. 
We assess the ability of frontier LLMs to generate RASP programs and find that this is a challenging task. With best-of-5 sampling and a 20-shot prompt, gpt-4-turbo-2024-04-09 correctly generates only 57 out of the 101 RASP programs in the test set. After adjusting for the difficulty of the programs, using the number of RASP operations as a proxy, the model achieves a normalized, weighted difficulty score of $0.29$ (the maximum score is 1.0). TracrBench aims to be a rich testbed for evaluating interpretability methods and accelerating their development.

\section{Method}\label{sec:method}

Current interpretability research faces challenges in rigorously evaluating novel methods due to the lack of models with fully understood internals. While Tracr compiles RASP programs into transformers with known mappings from weights to their functional form, writing programs in Tracr is time-consuming and difficult. This is partly because RASP is an unconventional, non-Turing-complete programming language that requires algorithms to be implemented differently than in standard Turing-complete languages like Python (see Appendix~\ref{example_program} for an example).

To address this issue, we propose to generate interpretability testbeds using LLMs, leveraging their ability to write code~\citep{achiam2023gpt, li2024humaneval}. We prompt LLMs to generate RASP programs that implement specified algorithms. We create TracrBench, a dataset of 121 RASP programs, by leveraging LLMs and manual annotation when they fail. These programs are then compiled into functional transformer weights using Tracr, resulting in transformer models with a known mapping between weights and their functional form. This allows researchers to validate the outputs of their novel interpretability methods against the ground truth. Our dataset of compiled models thus serves as an interpretability testbed. 

To generate a program, we condition a language model \(\mathcal{M}\) on a prompt \(\mathcal{P}\) that includes a description of the specific algorithm to be implemented and at least one example input-output pair (see Fig.~\ref{instruction_box}).
To optimize LLM performance, \(\mathcal{P}\) includes a detailed description of the RASP language and its five main components (\textsc{Select, Aggregate, SelectWidth, Map}, and \textsc{SequenceMap}), along with relevant Tracr source code defining these components and up to 20 RASP programs with their descriptions. We use Chain-of-thought prompting~\citep{wei2022chain} to encourage reasoning and planning before generating code (see Appendix~\ref{full_prompt} for the prompt). We create three variations of this prompt: Zero-Shot, One-Shot (extending Zero-Shot with an RASP program and its description), and 20-Shot (extending Zero-Shot with 20 RASP programs and their descriptions). 

Let $\mathcal{M}(\mathcal{P})$ represent the extracted program from the output of model $\mathcal{M}$ when conditioned on the prompt. We define a five-step verification pipeline to assess the correctness of the generated program $\mathcal{M}(\mathcal{P})$. Each step performs a specific verification relevant to the overall correctness of the program. Here are the five stages of the pipeline:

\begin{enumerate}\label{test_pipeline}
\item \textbf{Compilation and execution}: Test whether the program compiles without errors and runs error-free.

\item \textbf{Output correctness}: Test whether the function actually performs the correct computation and implements the specified program using 1,000 input-output pairs\footnote{The inputs are lists of random length between 1 and the maximum length selected for our compilation (which is 10).}
generated by a manually written Python function equivalent to the desired RASP program.
\item \textbf{Tracr validation}: Run the program through the inbuilt Tracr \citep{lindner2024tracr} validator\footnote{\href{https://github.com/google-deepmind/tracr/commit/9de2f8839ae3d4908c7c7c3b41912755fdf56011}{github.com}} to filter out certain programs that aren't converted to equivalent transformer weights.
\item \textbf{Transformer weights compilation}: Run the actual RASP-to-transformer compilation process to expose runtime errors like a division by zero.
\item \textbf{Compiled transformer correctness}: Empirically test whether the resulting transformer actually performs the same computation as the RASP code using the same 1,000 test input-output pairs from step 2.
\end{enumerate}

A program $\mathcal{M}(\mathcal{P})$ is considered correct if it passes all five steps; failure at any step counts as incorrect. This five-step verification pipeline helps identify and filter out programs with errors or inconsistencies, ensuring that the resulting dataset consists of high-quality, functionally equivalent RASP programs and transformer models. We employ best-of-5 sampling, allowing the model to attempt each task up to five times (from scratch) before moving on to the next. 
By evaluating the performance of LLMs with this process, we aim to assess the feasibility of using LLMs to create interpretability testbeds on demand.

\section{Dataset}\label{dataset}
Writing RASP code to generate Tracr interpretability test beds is labor-intensive and has a steep learning curve (see Appendix~\ref{example_program} for an example). This has impeded the adoption of Tracr as a method to evaluate novel interpretability methods. To address this issue, we present TracrBench, a novel dataset of Tracr models that enables interpretability researchers to quickly test methods on transformers with known mappings from weights to their functional form. The dataset is generated as follows. First, we select 121 simple, sequence-to-sequence algorithms that cover a diverse range of tasks and difficulty levels (see the full list in Appendix~\ref{algo_list}). We come up with these by sampling concrete algorithms from LLMs and manually selecting suitable ones. Some algorithms are also taken from~\citet{michaud2024opening}. We then prompt gpt-4-0125-preview (which was the most competitive model at the time) to generate a RASP program for each program description. We test all outputs with our verification pipeline and verify them manually, finding that 49 of the generated RASP programs are correct. We then manually write the remaining RASP programs, ensuring that all programs in the dataset are correct and of high quality. Finally, we take 20 samples to use as examples in the prompt and use the remaining 101 samples as our test set.

\begin{figure}[t]
    \centering
    \includegraphics[width=\linewidth]{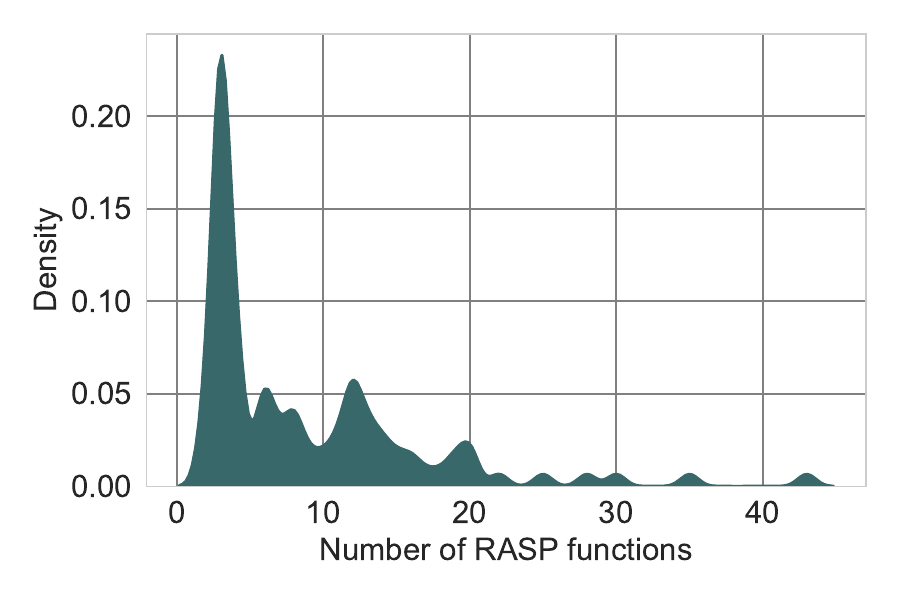}
    \vspace{-25pt}
    \caption{We show the distribution of RASP function calls within TracrBench using Kernel Density Estimation. The plot shows that most programs have around 6 RASP function calls, while a smaller number of more complex programs form a long tail.}
    \label{fig:difficulty_histogram}
\end{figure}

The resulting dataset contains RASP programs of various complexity, from simple elementwise operations to more complex programs that lead to transformers with 2.7 million parameters. We use the number of RASP functions (such as \texttt{Select} and \texttt{Aggregate}, but also \texttt{rasp.indices} and \texttt{rasp.tokens}) as a proxy for the difficulty of the algorithm. This approach is more accurate than counting lines of code because some programs may have many lines that don't involve RASP (see  Appendix~\ref{example_program} for an example). The distribution of task difficulties is depicted in Fig.~\ref{fig:difficulty_histogram} and Fig.~\ref{fig:programs_complexity}. The first figure shows that most programs are quite easy, containing 3 to 10 RASP functions, but there is a long tail of more complex programs with up to 43 RASP function calls. The second figure empirically depicts the success and failure of gpt-4-turbo on various programs, showing that the number of RASP functions is a better indicator of task complexity than the number of lines of code.

To facilitate the use of our dataset, we provide both the RASP programs and their corresponding compiled transformers as PyTorch~\citep{imambi2021pytorch} models in Transformerlens~\citep{nanda2022transformerlens}.

\begin{figure}
    \centering
\begin{tcolorbox}[breakable, enhanced, boxrule=0.5pt, left=2pt, right=2pt, top=0pt, bottom=0pt, colback=black!5!white, colframe=black!75!black, fontupper=\ttfamily\scriptsize]
\begin{lstlisting}[breaklines=true, basicstyle=\ttfamily\scriptsize]
# Your Task
Make a RASP program that replaces each element with the parity (0 for even, 1 for odd) of its index.
Example: [5, 5, 5, 5] --> [0, 1, 0, 1]
\end{lstlisting}
\end{tcolorbox}
\caption{The description of the target algorithm to implement that is part of the prompt for the LLM.}
\label{instruction_box}
\end{figure}

\section{Experiments}
In this section, we evaluate the capability of LLMs to generate correct RASP programs. As described in Section~\ref{sec:method}, we condition an LLM on a prompt that includes a program specification, a detailed description of the RASP language, and important parts of the RASP source code. We use three variations of the prompt: a zero-shot, a one-shot prompt, and a 20-shot prompt. These different prompt variations are used to assess how including examples affects the LLM's performance in generating RASP programs. 

We evaluate the generated RASP programs using the verification pipeline described in Section~\ref{test_pipeline}. A program is considered correctly implemented if it passes all five pipeline steps. To account for the variance in program difficulty, we introduce a second metric called the difficulty-weighted score, which weights each success by the number of RASP functions in the program. Summing these weighted scores across tasks provides us with a composite score that more effectively represents the model's proficiency.

We first evaluate the performance of different prompts using gpt-3.5-turbo-0125 and gpt-4-turbo-2024-04-09. To minimize compute costs, we evaluate the performance of additional models using only the full prompt. These models include gpt-4o-2024-05-13, claude-3-haiku-20240307, claude-3-sonnet-20240229, claude-3-opus-20240229 and claude-3-5-sonnet-20240620. All models are evaluated on the test set with 101 samples and sampled at temperature 0.9, with top-p=$0.95$. To distinguish between an LLM's general programming ability and its RASP-specific capabilities, we establish a baseline where the LLM writes a Python program for the same target algorithms.

\subsection{Results}
\begin{figure}[t]
    \centering    \includegraphics[width=\linewidth]{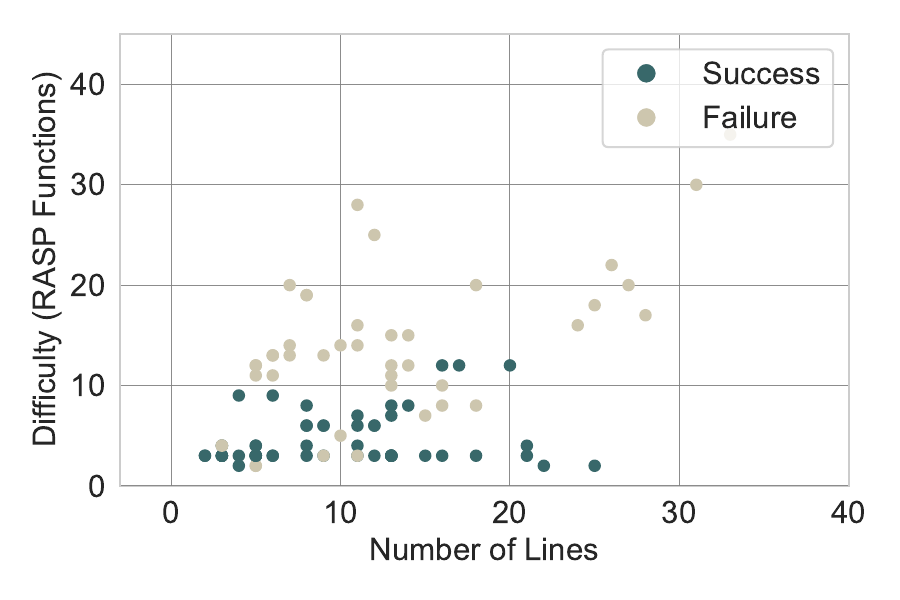}
    \vspace{-25pt}
    \caption{We compare the number of RASP functions and program lines as proxies for task difficulty. When plotting the pass-rate of gpt-4o-turbo on all programs, we can see that the number of RASP functions is a better indicator of task complexity than the total lines of code.}
    \label{fig:programs_complexity}
\end{figure}

The results of our experiment, visualized in Fig.~\ref{fig:combined}, show that state-of-the-art LLMs are able to understand the RASP language and, to some extent, generate correct RASP programs. Adding examples to the prompt clearly improves the performance, as shown with gpt-4-turbo and gpt-3.5-turbo.
Overall, gpt-4-turbo achieves the highest pass rate of 56\%, outperforming claude-3-opus with a pass rate of 46\%. In comparison, when generating Python programs for the target algorithms, gpt-4-turbo achieves a pass rate of 96\%. 
When taking the difficulty of the target algorithms into account, i.e., when using the difficulty-weighted score as a metric, we observe that the successes are strongly concentrated among the easy, low-difficulty programs (see Fig.~\ref{fig:programs_complexity}) with gpt-4-turbo achieving a score of $0.29$ (out of $1.0$) and gpt-4o performing best with a score of $0.31$. Claude-3-5-sonnet has a similar pass rate ($0.45$) to claude-3-opus ($0.46$), however, it achieves a higher difficulty-weighted score
($0.27$), than claude-3-opus ($0.23$).

These results suggest that frontier LLMs cannot yet competently generate correct RASP programs.
The relatively poor performance of generating RASP programs compared to conventional programming languages like Python may be attributed to RASP's limited representation in LLM training data. This finding highlights that the ability of frontier LLMs to extend their reasoning and programming capabilities to low-resource programming languages is limited, which may stand in contrast with their generalization in natural low-resource languages ~\citep{reid2024gemini}.

\section{Related Work}
Evaluating novel interpretability methods is challenging~\citep{eis_vii}. While previous work has addressed this issue, it mainly focused on input-output level interpretability \citep{casper2024satml, casper2023red, mazeika2022trojan}, human evaluations~\citep{templeton2024scaling},
or disentangling attributes of different entities~\citep{huang2024ravel}.
RASP~\citep{weiss2021thinking} a programming language computationally equivalent to transformer, and Tracr~\citep{lindner2024tracr}, which compiles RASP programs into corresponding transformers, have been used to create interpretable models for validating interpretability methods \citep{conmy2023towards}. However, writing RASP programs in sufficient quantity is very time-consuming, which hinders the broad adoption of Tracr to evaluate interpretability methods. Notably, Tracr weights are more sparse and simple than any set of weights likely to result from gradient descent. Therefore, a method capable of interpreting Tracr weights may not necessarily be able to interpret trained transformers. However, interpretability methods that are capable of interpreting trained transformers should also be capable of interpreting Tracr transformers. Thus the latter still serve as a valid method to test (but not to develop) useful interpretability methods.
Finally, both \citet{thurnherr2024decompiling} and \citet{langosco2024meta} programmatically generate large quantities of RASP programs with their corresponding weights to train decompiler models that generate RASP programs for a given set of transformer weights. Their RASP programs are, however, randomly generated by re-combining a few elemental operations, which leads to models that are often hard to decipher and do not correspond to realistic algorithms.

LLMs have been explored for generating datasets for model evaluations~\citep{perez2022discovering} and automating part of the interpretability workflow~\citep{bills2023language}. We extend this by using LLMs to scalably generate realistic and interpretable RASP programs. The generated programs serve as a test bed for evaluating interpretability methods.

\section{Conclusion}
We demonstrate that LLMs can be used to generate interpretability test beds. However, their performance rapidly deteriorates with the increasing difficulty of RASP programs, indicating that frontier LLMs struggle to generate interpretability test beds at scale. We expect that these current limitations, likely due to Tracr's low-resource nature, will diminish as LLM capabilities continue to advance. Finally, we introduce TracrBench, a novel dataset comprising 121 transformers with known mappings from weights to functional form. Its intended use is the testing of interpretability methods. It is unsuitable as a target for interpretability method development due to its small size and the fact that Tracr weights are very dissimilar to those of trained transformers in terms of sparsity and matrix-rank.
TracrBench serves as a valuable resource for evaluating and comparing interpretability methods, facilitating the development of more effective techniques for understanding the inner workings of transformer-based models. 

\section{Author Contributions}
\textbf{Hannes Thurnherr} executed the whole project, developed the prompts, created the dataset (i.e., the Tracr programs) with the help of LLMs and manual labour,w ran all experiments and wrote the paper. \textbf{Jérémy Scheurer} developed the idea and ran exploratory experiments, oversaw the project, including detailed guidance on directions, experiments, presentation, and the final paper.
\newpage
\bibliography{references}
\bibliographystyle{icml2024}

\newpage
\appendix
\onecolumn
\appendix

\section{Complete list of Algorithms}\label{algo_list}

\begin{center}
\begin{longtable}{|p{0.3\textwidth}|p{0.65\textwidth}|}
\hline
\textbf{Program Name} & \textbf{Program Description} \\ 
\hline
make\_sum\_digits & replaces each element with the sum of its digits. \\ \hline
make\_absolute & takes the absolute value of each element in the sequence. \\ \hline
make\_first\_element & returns the first element of the sequence. \\ \hline
make\_nth\_fibonacci & replaces each element with the nth Fibonacci number. \\ \hline
make\_count\_greater\_than & replaces each element with the number of elements greater than it in the sequence. \\ \hline
make\_double\_first\_half & doubles the first half of the sequence. For uneven number of entries, round up to half. \\ \hline
make\_decrement & decrements each element in the sequence by 1. \\ \hline
make\_count\_frequency & counts the frequency of each unique element. \\ \hline
make\_increment\_by\_index & increments each element by its index. \\ \hline
make\_decrement\_to\_multiple\_of\_three & decrements each element until it becomes a multiple of 3. \\ \hline
make\_hyperbolic\_cosine & applies the hyperbolic cosine to each element. \\ \hline
make\_check\_fibonacci & checks if each element is a Fibonacci number. \\ \hline
make\_square\_root & takes the square root of each element. \\ \hline
make\_increment\_odd\_indices & increments elements at odd indices. \\ \hline
make\_hyperbolic\_tangent & applies the hyperbolic tangent to each element. \\ \hline
make\_hyperbolic\_sine & applies the hyperbolic sine to each element. \\ \hline
make\_zero\_every\_third & sets every third element to zero. \\ \hline
make\_element\_second & replaces each element with the second element of the sequence. If the sequence has fewer than two elements you should return [None]. \\ \hline
make\_mirror\_first\_half & mirrors the first half of the sequence to the second half. \\ \hline
make\_sorting & sorts the sequence. \\ \hline
make\_increment & increments each element in the sequence by 1. \\ \hline
make\_rank & ranks each element according to its size. \\ \hline
make\_factorial & replaces each element with its factorial. \\ \hline
make\_count\_less\_than & replaces each element with the number of elements less than it in the sequence. \\ \hline
make\_cube\_each\_element & cubes each element in the sequence. \\ \hline
make\_cube\_root & takes the cube root of each element. \\ \hline
make\_round & rounds each element to the nearest integer. \\ \hline
make\_multiply\_by\_length & multiplies each element by the number of elements in the sequence. \\ \hline
make\_increment\_to\_multiple\_of\_three & increments each element until it becomes a multiple of 3. \\ \hline
make\_sign & determines the sign of each element (positive, negative, or zero). \\ \hline
make\_cosine & applies the cosine function to each element. \\ \hline
make\_divide\_by\_length & divides each element by the number of elements in the sequence. \\ \hline
make\_negation & negates each element in the sequence. \\ \hline
make\_sine & applies the sine function to each element. \\ \hline
make\_histogram & creates a histogram of elements. \\ \hline
make\_element\_double & doubles each element in the sequence. \\ \hline
make\_zero\_even\_indices & sets all even indices to zero. \\ \hline
make\_tangent & applies the tangent function to each element. \\ \hline
make\_count\_occurrences & replaces each element with the number of times it appears in the sequence. \\ \hline
make\_compute\_median & computes the median of the sequence. \\ \hline
make\_halve\_second\_half & halves the second half of the sequence. Note that you should divide sequences with odd number of elements into [first half of size n, second half of size n+1]. \\ \hline
make\_triple & triples each element in the sequence. \\ \hline
make\_arctangent & applies the arctangent function to each element. \\ \hline
make\_square\_each\_element & squares each element in the sequence. \\ \hline
make\_check\_power\_of\_n & checks if each element is a power of n (make the default for n 2). 1 and n itself, also count as power of $n$ since they correspond to $n^0$ and $n^1$. \\ \hline
make\_binarize & binarizes elements based on a threshold (make the default threshold 3). \\ \hline
make\_average\_first\_last & sets each element to the average of the first and last elements. \\ \hline
make\_check\_increasing & checks if every element is greater than or equal to the previous one. The output should only contain all ones if every entry, that has a previous entry, meets this condition. Otherwise the output should be all 0s. \\ \hline
make\_identity & returns the same sequence. \\ \hline
make\_apply\_threshold & applies a threshold, setting elements below it to zero (make the default threshold 3). \\ \hline
make\_replace\_small\_tokens & replaces tokens smaller than a threshold with zero (make the default threshold 2). \\ \hline
make\_swap\_odd\_index & swaps the $n$th with the $n+1$th element if $n\%2==1$. Note that this means that the first element will remain unchanged. The second will be swapped with the third and so on. \\ \hline

make\_check\_descending & checks if the sequence is in descending order. \\ \hline
make\_rotate\_left & rotates elements to the left by 1 position. \\ \hline
make\_remove\_duplicates & removes (replaces with 0) duplicates from the sequence. The first occurrences of the duplicated numbers also have to be removed. \\ \hline
make\_scale\_by\_max & scales each element by the maximum value in the sequence. \\ \hline
make\_sum\_with\_next & replaces each element with the sum of it and the next element. For the last element you can sum it with itself. \\ \hline
make\_swap\_elements & swaps two elements at specified indices (make the default indices 0 and 1). If an input sequence only has 1 element return [None]. \\ \hline
make\_one\_if\_equal\_to\_next & sets elements to one if they are equal to the next element. The last element should be compared with the first. \\ \hline
make\_swap\_consecutive & swaps every two consecutive elements. If the number of entries is odd, the last entry should stay in place. \\ \hline
make\_check\_palindrome & checks if the sequence is a palindrome. \\ \hline
make\_next\_prime & replaces each element with the next larger prime number. If the element is already prime, it should stay the same. \\ \hline
make\_mask\_sequence & masks a sequence, replacing every element with 0 except the one at a specified index (make the default index 1). \\ \hline
make\_wrap & wraps each element within a range (make the default range [2, 7]). Wrapping here means that the values are projected into the range starting from the lower bound, once they grow larger than the upper bound, they start again at the lower. \\ \hline
make\_alternate\_elements & alternates elements with their indices. \\ \hline
make\_check\_last\_two\_equal & checks whether the last two entries of a sequence are equal. If the sequence only has one entrance, return [0]. \\ \hline
make\_insert\_zeros & inserts zeros between each element. This means that the latter half of the sequence will be cut off (no 4 and 5 in the following example). \\ \hline
make\_last\_element & returns the last element of the sequence and pads the rest with zeros. \\ \hline
make\_difference\_to\_next & replaces each element with the difference to the next element. \\ \hline
make\_invert\_if\_sorted & inverts the sequence if it is sorted in ascending order, otherwise leaves it unchanged. \\ \hline
make\_logarithm & applies logarithm base 10 to each element. \\ \hline
make\_product\_with\_next & replaces each element with the product of it and the next element. The last element should be multiplied with itself. \\ \hline
make\_check\_multiple\_of\_first & checks if each element is a multiple of the first element. \\ \hline
make\_sum\_of\_last\_two & returns the sum of the last two elements in the sequence. If the sequence only has one entry, return [None]. \\ \hline
make\_pairwise\_sum & replaces each element with the sum of it and the previous element. The first element can be left as it is. \\ \hline
make\_polynomial & evaluates a polynomial with sequence elements as parameters. The x is represented by the first entry, the rest are parameters for example $$[3,4,2,1]$$ is equal to $4x^2+2x+1$ for $x =3 so 4*3^2 + 2*3 + 1 = 36 + 6 +1 = 43$ represented as $$[43, 43, 43, 43]$$. \\ \hline
make\_flip\_halves & flips the order of the first and second half of the sequence. Note that you should divide sequences with odd number of elements into [first half of size n, second half of size n+1]. \\ \hline
make\_arcsine & applies the arcsine function to each element. \\ \hline
make\_check\_divisibility & checks if the sequence consists of numbers divisible by some parameter (make the default 3). \\ \hline
make\_arccosine & applies the arccosine function to each element. \\ \hline
make\_check\_all\_equal & checks whether all elements are equal. \\ \hline
make\_position & replaces each element with its position in the sequence. \\ \hline
make\_set\_to\_median & replaces each element with the median of all elements. \\ \hline
make\_swap\_min\_max & swaps the largest and smallest elements in the sequence. If the maximum or minimum appears more than once, both occurrences must be replaced. \\ \hline
make\_clip & clips each element to be within a range (make the default range [2, 7]). "Clipping" means that values outside of the range, are turned into the lower or upper bound, whichever is closer. \\ \hline
make\_pairwise\_max & makes each element the maximum of it and the previous element, leaving the first element as it is. \\ \hline
make\_check\_alternating & checks if the sequence consists of alternating odd and even numbers. If this is not true, all the entries in the output sequence should be zero. \\ \hline
make\_exponential & exponentiates each element. \\ \hline
make\_interleave\_reverse & interleaves elements with their reverse order Numbers at the odd indices should be in reverse order. \\ \hline
make\_element\_divide & divides each element by the division of the first two elements. If either the first or second element are zero, or if the sequence has fewer than two entries, you should just return the original sequence. \\ \hline
make\_set\_to\_index & sets elements to their index value. \\ \hline
make\_check\_multiple\_of\_n & checks if all elements are a multiple of n (set the default at 2). The output should be all 1s if this is true for all elements, otherwise all 0s. \\ \hline
make\_swap\_first\_last & swaps the first and last elements of the sequence. If the sequence only has one entry, just return the original sequence. \\ \hline
make\_test\_at\_least\_two\_equal & checks whether at least two elements are equal. \\ \hline
make\_reflect & reflects each element within a range (make the default range [2, 7]). Reflect means that the values will be projected into the range, "bouncing" from the borders, until they have travelled as far in the range as they travelled outside of it. \\ \hline
make\_check\_square & checks for every entry of the sequence whether it is a square number or not. \\ \hline
make\_count\_prime\_factors & replaces each element with the number of prime factors it has. \\ \hline
make\_zero\_if\_less\_than\_previous & sets elements to zero if they are less than the previous element. \\ \hline
make\_element\_subtract\_constant & subtracts a constant from each element (make the default constant 2). \\ \hline
make\_check\_prime & checks if each element is a prime number. \\ \hline
make\_index\_parity & replaces each element with the parity (0 for even, 1 for odd) of its index. \\ \hline
\end{longtable}
\end{center}

\clearpage
\section{Example Program}\label{example_program}

RASP, a programming language designed to be computationally equivalent to transformers, requires a conceptually different approach to implementing algorithms compared to conventional programming languages. For instance, sorting algorithms in RASP must be implemented unconventionally due to the language's unique constraints.
Unlike traditional programming languages that allow iteration over a sequence, RASP processes all elements in a sequence in parallel, mimicking the behavior of transformers. Consequently, a sorting algorithm in RASP would count, for each entry, the number of other entries smaller than itself and then use these counts to rearrange the original elements.
While this approach would be considered inefficient in conventional programming languages, it is a straightforward implementation under the constraints of RASP. This example highlights the need for a different mindset when writing algorithms in RASP, as the language's parallel processing nature requires unconventional solutions to common problems.

\lstset{
    language=Python,
    basicstyle=\ttfamily\footnotesize,
    keywordstyle=\color{blue},
    commentstyle=\color{gray},
    stringstyle=\color{red},
    showstringspaces=false,
    numbers=left,
    numberstyle=\tiny\color{gray},
    breaklines=true,
    frame=single,
    tabsize=4
}
\begin{figure}[ht]
\begin{lstlisting}
def make_sort_unique(vals: rasp.SOp, keys: rasp.SOp) -> rasp.SOp:
    smaller = rasp.Select(keys, keys, rasp.Comparison.LT) # find the smaller elements for each entry
    target_pos = rasp.SelectorWidth(smaller) # count the number of smaller elements for each entry
    sel_new = rasp.Select(target_pos, rasp.indices, rasp.Comparison.EQ) # create the rearrangement selector according to target_pos
    return rasp.Aggregate(sel_new, vals) # apply the rearrangement selector to the original sequence

def make_sort(vals: rasp.SOp, keys: rasp.SOp, *, max_seq_len: int, min_key: float) -> rasp.SOp:
    keys = rasp.SequenceMap(lambda x, i: x + min_key * i / max_seq_len, keys, rasp.indices) # turn all the elements unique by adding a small fraction of their index
    return make_sort_unique(vals, keys) # apply sort_unique to the sequence using the now unique elements as keys
\end{lstlisting}
\end{figure}

RASP programs written for the Tracr compiler are written in Python using the tracr.rasp module. Sometimes they consist of a number of lines where the tracr.rasp module is not used. These parts of the RASP program can be written independently of one's understanding of the RASP language.
The following is an example of a program where most lines don't involve RASP. This illustrates why the number of rasp functions in a program is a better approximation of difficulty than the number of total lines when it comes to evaluating a model's ability to write RASP code.

\begin{figure}[ht]
\begin{lstlisting}
def primecheck(n):
    for i in range(2,int(n/2)):
        if n%i==0:
            return 0
    return 1

def make_check_prime() -> rasp.SOp:
    return rasp.Map(lambda x: primecheck(x), rasp.tokens)
\end{lstlisting}
\end{figure}

\section{Full Prompt}
\label{full_prompt}
\lstset{
    language=,
    basicstyle=\ttfamily\footnotesize,
    keywordstyle=\color{black},
    commentstyle=\color{gray},
    stringstyle=\color{black},
    showstringspaces=false,
    numbers=left,
    numberstyle=\tiny\color{gray},
    breaklines=true,
    frame=single,
    tabsize=4
}
\begin{tcolorbox}[breakable, pad at break*=3mm, left=7mm, right=5mm, top = 3mm, bottom = 3mm, colback=black!5!white, colframe=black!75!black, title={Prompt "Paraphrased + Tip about different stock"}, fontupper=\small\ttfamily]
\begin{lstlisting}[breaklines=true, basicstyle=\ttfamily\footnotesize]

# Introduction to Task:
Your assignment is to generate RASP programs capable of implementing a variety of algorithms using sequence operations. "RASP" stands for "Restricted Access Sequence Processing Language". RASP allows you to articulate complex sequence to sequence in a format equivalent to what a neural network of the transformer architecture can do. RASP programs always output a sequence that has the same length as the input sequence.

# Your Task
Make a RASP program that replaces each element with the parity (0 for even, 1 for odd) of its index.
Example: [5, 5, 5, 5] --> [0, 1, 0, 1]
Name the function that you can call to make this program 'make_index_parity()'

Keep your task in mind while reading the following information.

# Understanding RASP:

RASP programs are unique because they always process sequences and output transformed sequences of equivalent length. While doing so they void conditional branches or loops if possible. Instead, they rely on a series of operations that interpret and manipulate the input data in a sequence-to-sequence fashion. The length of the sequence never changes during this process.

## Fundamental Principles:

- Input and Output: Each RASP program receives an input sequence and yields an output sequence of identical length.
- Structure: Loops and if statements cannot depend on attributes or individual elements of the input sequence. If you make loops, they should have a fixed length or depend on a "max_sequence_length" parameter.
- Operation Calls: Programs can only invoke core RASP functions or refer to other RASP programs. Never attempt to access the internals of the sequence.

## Technical operational Jargon:

Here are descriptions of various operations that are used in RASP.

- `rasp.Select`: Matches elements from two sequences based on a boolean comparison condition and returns a corresponding matrix of "True" and "False" values called a selector.
- `rasp.Aggregate`: takes as input a selector and an SOp (Sequence Operation, which is an operation that transforms a sequence), and produces an SOp that averages the value of the SOp weighted by the selection matrix.
- `rasp.Map`: Transforms a sequence by applying a function to each element
- `rasp.SequenceMap`: Produces a new sequence based on two previous sequences and a lambda function that gets applied to each pair of elements.
- `rasp.SelectorWidth`: returns the number of "True" values in each row of a selector

### Function overview:

#### Select:
Function: Creates a selector to define relationships between elements of sequences.
Syntax: `rasp.Select(keys: SOp, queries: SOp, predicate: Predicate)`
Example: `rasp.Select(rasp.indices, rasp.indices, rasp.Comparison.EQ)` selects elements where indices are equal.

#### Aggregate:
Function: Takes as input a selector and an SOp, and produces an SOp that averages the value of the SOp weighted by the selection matrix.
Syntax: `rasp.Aggregate(selector: Selector, sop: SOp, default: Optional[VT] = None)`
Example: `rasp.Aggregate(select_all, any_negative, default=0)` aggregates based on select_all.

#### Map:
Function: Applies a function element-wise on the input SOp.
Syntax: `(f: Callable[[Value], Value], inner: SOp)`
Example: `Map(lambda x: x + 1, tokens)` adds 1 to each element of tokens.

#### SequenceMap:
Function: Applies a function element-wise on two given SOps.
Syntax: `rasp.SequenceMap(f: Callable[[Value, Value], Value], fst: SOp, snd: SOp)`
Example: `rasp.SequenceMap(lambda x, y: x - y, rasp.indices, rasp.tokens)` subtracts tokens from indices.

#### SelectorWidth:
Function: Returns the "width" of a selector, which corresponds to the number of "True"-values in each row.
Syntax: `rasp.SelectorWidth(selector: Selector)`
Example: `rasp.SelectorWidth(selectAll)`

#### Tokens, Indices:
rasp.tokens: The original input sequence.
rasp.indices: Returns the position index at each token.

### Example use of above Functions:
This is an example use the rasp.Select function. Here, it produces a selector based on rasp.tokens applied to itself with the "Greater Than" or GT comparison operator:

```python
greater_than_selector = rasp.Select(rasp.tokens, rasp.tokens, rasp.Comparison.GT).named("greater_than_selector")
```
If the rasp.tokens-sequence is [1, 2, 3, 4] the selector will look like this:
[False, True, True, True]
[False, False, True, True]
[False, False, False, True]
[False, False, False, False]
If we now apply this to the original rasp.tokens again with:
```python
output = rasp.Aggregate(greater_than_selector, rasp.tokens)
```
We will get an average of all the values selected in each row. The output looks like this:
[3, 3.5, 4, None]
[
3, # as an average of the selected 2,3 and 4
3.5, # as an average of the selected 3 and 4
4, # as an average of the selected 4
None # because none of the values were selected as none of them are greater than 4 at this position. So, None, which is always the default value, takes this spot.
]
Note that, in the programs you create, you should avoid using rasp.Aggregate with selectors that have more than one true value in each row. In other words: you can use rasp.Aggregate to shift elements around, but avoid using it for averaging multiple elements. However, using rasp.SelectWidth with selectors that have more than one "True" value per row is completely fine.
If we now call:
```python
count_GT_selector = rasp.SelectorWidth(greater_than_selector)
```
We will get a sequence that contains the count of the truth values in each row:
[3,2,1,0]
If we call:
```python
map_count_GT = rasp.Map(lambda x: x*3+1, count_GT_selector)
```
We will get a sequence where this lambda function has been applied to all the values of count_GT_selector:
[10, 7, 4, 1]

But if we call:
```python
sequenceMap_combination = rasp.SequenceMap(lambda x, y: x*y+x, count_GT_selector, output)
```
We get an output where the sequences "count_GT_selector" and "output" are combined element-wise according to the lambda function.
At this point, "count_GT_selector" is [3,2,1,0] and output is [3, 3.5, 4, None], so sequenceMap_combination is [12, 9, 5, None]
[
12, #because 3 * 3 + 3 = 12
9, #because 2 * 3.5 + 2 = 9
5, #because 1 * 4 + 1 = 5
0 #because 0 * None + 0 = 0
]

# Rules and Constraints:
- Use provided operation types (Select, Aggregate, SelectorWidth Map, SequenceMap) as the building blocks of your program. Feel free to be creative in how to combine them but remember which kind of output (Selector or Sop) they produce.
- Each operation must be traceable and reproducible, implying a transparent translation from instructions to action.

# Source Code
To make you better understand the RASP language you can look at the following code. These are the most important parts of rasp.py, which defines the library of RASP. Use this as a reference to find out what kind of functions exist in RASP, which inputs they take, and what they do.



# Example use of Functions:
This is an example use the rasp.Select function. Here, it produces a selector based on rasp.tokens applied to itself with the "Greater Than" or GT comparison operator:

```python
greater_than_selector = rasp.Select(rasp.tokens, rasp.tokens, rasp.Comparison.GT).named("greater_than_selector")
```
If the rasp.tokens-sequence is [1, 2, 3, 4] the selector will look like this:
[False, True, True, True]
[False, False, True, True]
[False, False, False, True]
[False, False, False, False]
If we now apply this to the original rasp.tokens again with:
```python
output = rasp.Aggregate(greater_than_selector, rasp.tokens)
```
We will get an average of all the values selected in each row. The output looks like this:
[3, 3.5, 4, None]
[
3, # as an average of the selected 2,3 and 4
3.5, # as an average of the selected 3 and 4
4, # as an average of the selected 4
None # because none of the values were selected as none of them are greater than 4 at this position. So, None, which is always the default value, takes this spot.
]
Note that, in the programs you create, you should avoid using rasp.Aggregate with selectors that have more than one true value in each row. In other words: you can use rasp.Aggregate to shift elements around, but avoid using it for averaging multiple elements. However, using rasp.SelectWidth with selectors that have more than one "True" value per row is completely fine.
If we now call:
```python
count_GT_selector = rasp.SelectorWidth(greater_than_selector)
```
We will get a sequence that contains the count of the truth values in each row:
[3,2,1,0]
If we call:
```python
map_count_GT = rasp.Map(lambda x: x*3+1, count_GT_selector)
```
We will get a sequence where this lambda function has been applied to all the values of count_GT_selector:
[10, 7, 4, 1]

But if we call:
```python
sequenceMap_combination = rasp.SequenceMap(lambda x, y: x*y+x, count_GT_selector, output)
```
We get an output where the sequences "count_GT_selector" and "output" are combined element-wise according to the lambda function.
At this point, "count_GT_selector" is [3,2,1,0] and output is [3, 3.5, 4, None], so sequenceMap_combination is [12, 9, 5, None]
[
12, #because 3 * 3 + 3 = 12
9, #because 2 * 3.5 + 2 = 9
5, #because 1 * 4 + 1 = 5
0 #because 0 * None + 0 = 0
]


Start your process by looking at the examples and the RASP language basics, then write down a plan based on the information in the files and the examples above, and then write your program.
If your plan includes the usage of a certain function, look up all of the allowed parameters for this function, write them down before you start writing the program and make sure you do not make up any new parameters to any of the RASP functions.
Note that you are not allowed to directly call the above examples as functions in your code, without explicitly writing/copying them into your output yourself. This means if you want to call functions like `make_length()` or `shift_by()`, you have to rewrite them in your output code.

# Output Format
Use the following Format for your answer:

<Task>
[Reiterate your understanding of the task and add a new example of an input and the corresponding desired output.]
</Task>

<Plan>
[Your plan on how the program should broadly work.]
[Some details on which functions you'll have to use and what their inputs will be.]
</Plan>

<PlanVerification>
[Look back at your plan. Will it really work? Is this compatible with the functionality of the functions you're using? Are you using your functions correctly? (Look at the source code to verify this) Answer these questions here explicitly]
[List changes you have to make to the plan based on your verification]
</PlanVerification>

```python
[write out your RASP-python code in a code block here]
```

### Example Use of Format:

Here is an example of how you might use this output format:

<Task>
The task is to create a RASP program that takes a sequence and returns a new sequence of identical length where each element is the maximum value found in the original sequence.

For example:
max = make_max()
max([1,2,6,-2,1]) # returns [6,6,6,6,6]
</Task>

<Plan>
1. Create a selector that compares each element with every other element using a "Less Than or Equal" (LEQ) comparison.
2. Use SelectorWidth to count the number of elements that each element is less than or equal to.
3. The maximum element will have a count equal to the length of the sequence, so create a selector that selects the elements where the count from step 2 equals the length of the sequence.
4. Use Aggregate with the selector from step 3 to broadcast the maximum element across the entire sequence.

The functions we will use include:
- Select: for creating the comparison selector.
- SelectorWidth: for counting the number of comparisons that are true for each element.
- Map: for creating a sequence of the sequence length.
- Aggregate: for selecting and broadcasting the maximum element.
</Plan>

<PlanVerification>
The plan seems feasible and aligns with the capabilities of the RASP functions:
- The Select operation can create a comparison matrix that identifies where each element is less than or equal to every other element.
- SelectorWidth can count the number of True comparisons for each element.
- Map can create a sequence where each element is the length of the sequence.
- Aggregate can then broadcast the maximum element where the comparison count equals the sequence length.

There are no changes needed for the plan based on verification.
</PlanVerification>

```python
def make_max() -> rasp.SOp:
    # Selector that creates a comparison matrix where each element is compared to every other element.
    leq_selector = rasp.Select(rasp.tokens, rasp.tokens, rasp.Comparison.LEQ).named("leq_selector")

    # Count the number of comparisons where each element is less than or equal to other elements.
    leq_count = rasp.SelectorWidth(leq_selector).named("leq_count")

    # Create a Map to get the sequence length for each element.
    sequence_length = rasp.Map(lambda x: len(x), rasp.tokens).named("sequence_length")

    # Selector that selects the element where the leq_count equals the sequence_length.
    max_element_selector = rasp.Select(leq_count, sequence_length, rasp.Comparison.EQ).named("max_element_selector")

    # Use Aggregate to broadcast the maximum element across the entire sequence.
    max_sequence = rasp.Aggregate(max_element_selector, rasp.tokens).named("max_sequence")

    return max_sequence
```


# Your Task
Make a RASP program that replaces each element with the parity (0 for even, 1 for odd) of its index.
Example: [5, 5, 5, 5] --> [0, 1, 0, 1]
Name the function that you can call to make this program 'make_index_parity()'


Examples provided are references; use them to grasp the syntax and structure required for RASP. From there, your original programs should follow these established patterns but are not limited to the examples' specific functions.

Keep in mind:
- Adhere strictly to RASP's core operations.
- Keep your programs simple, if possible. (E.g. For identity, just return rasp.Map(lambda x: x, rasp.tokens)
- Meticulously add comments to your code for clarity.
- Output functional, executable Python code utilizing RASP's parameters.
- Don't import any additional packages. Write pure RASP code.
- Provide functional, complete Python code, not pseudo-code or placeholders.

Also Note:
- Do not import rasp. It is already imported. You should also not try to import the rasp components individually.
- Aggregate functions should always have None as the default (meaning you should leave the default as is.) This is because we want to compile these functions later, which only works with a default of None.
- Again, do not use any functions from the example without defining them yourself. You cannot assume any function from the examples is already defined.
- If your `make_x()` functions have additional parameters like `make_x(n)` or `make_x(threshold)`, you should always have a default value like `make_x(threshold = 2)`
- Avoid the `rasp.Full()` functionality. It will prevent compiling. Instead of `rasp.Full(n)`` use the following function: `rasp.Map(lambda x: n, rasp.indices)`

Endeavour to follow these guidelines to construct accurate and efficient RASP programs. Your expertise in Python will be fundamental to this task, so make sure that your code is both clean and precise, adhering to the RASP principles.

\end{lstlisting}
\end{tcolorbox}

\end{document}